\pdfoutput=1
\documentclass[11pt]{article}
\usepackage[final]{acl}
\usepackage{times}
\usepackage{latexsym}
\usepackage{wrapfig}
\usepackage{booktabs}
\usepackage[T1]{fontenc}
\usepackage[utf8]{inputenc}
\usepackage{amsmath}
\usepackage{microtype}
\usepackage{amssymb} 
\usepackage{inconsolata}
\usepackage{graphicx}
\usepackage{bbm}

\title{Domain Gating Ensemble Networks for AI-Generated Text Detection}

\author{Arihant Tripathi$^*$, \hspace{0.25cm} Liam Dugan$^*$, \hspace{0.25cm} Charis Gao, \hspace{0.25cm} Maggie Huan,\\
\textbf{Emma Jin}, \hspace{0.25cm} \textbf{Peter Zhang}, \hspace{0.25cm} \textbf{David Zhang}, \hspace{0.25cm} \textbf{Julia Zhao}, \hspace{0.25cm} \textbf{Chris Callison-Burch}\\
University of Pennsylvania\\\ 
{\tt \normalsize \{atrip, ldugan, ccb\}@seas.upenn.edu}\\ \tt \normalsize }

\begin{document}
\maketitle
\begingroup
\renewcommand\thefootnote{*}
\footnotetext{Equal contribution}
\endgroup

\begin{abstract}
As state-of-the-art language models continue to improve, the need for robust detection of machine-generated text becomes increasingly critical. However, current state-of-the-art machine text detectors struggle to adapt to new unseen domains and generative models. In this paper we present DoGEN (Domain Gating Ensemble Networks), a technique that allows detectors to adapt to unseen domains by ensembling a set of domain expert detector models using weights from a domain classifier. We test DoGEN on a wide variety of domains from leading benchmarks and find that it achieves state-of-the-art performance on in-domain detection while outperforming models twice its size on out-of-domain detection. We release our code and trained models\footnote{\url{https://github.com/Siris2314/dogen}} to assist in future research in domain-adaptive AI detection.
\end{abstract}

\section{Introduction}
Large Language Models (LLMs) are extremely good at generating human-like text. Such text is frequently mistaken for human-written text by non-expert readers and scholars alike \cite{dugan-etal-2020-roft,dugan-etal-2023-roft,clark-etal-2021-thats}. Therefore, robust and cheap AI detection technology would help significantly reduce the plethora of negative externalities of LLMs such as scientific fraud \cite{lund2023chatgpt}, targeted phishing scams \cite{hazell2023large}, and disinformation \cite{spitale-etal-2023-disinformation}.

\begin{figure}
    \centering
    \includegraphics[width=\columnwidth]{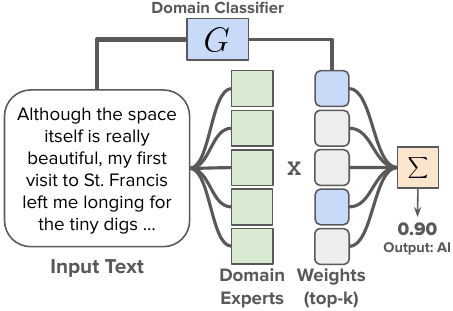}
    \caption{The Domain Gating network splits the input into a probability distribution over $N$ experts. The output is a weighted sum of the outputs of the top $k$ experts.}
    \label{fig:pg1}
\end{figure}

Recent work in AI text detection has largely focused on training single models on a variety of domains and generators \cite{emi2024technicalreportpangramaigenerated, hu2023radarrobustaitextdetection}. However, these models have been shown to lack robustness to unseen domains and generators \cite{dugan-etal-2024-raid}.

To address these limitations, we propose DoGEN (Domain Gating Ensemble Networks). We first use a \textit{domain router} network to output a probability distribution over $N$ domains for a given input document. This distribution is then used to weight an ensemble of domain-specific expert detectors in order to predict the correct output class (Figure \ref{fig:pg1}). In our experiments we show that DoGEN achieves state-of-the-art performance for in-domain detection and also outperforms other similar ensembling techniques on unseen domains and generators. 

\begin{figure*}
    \centering
    \includegraphics[width=2\columnwidth]{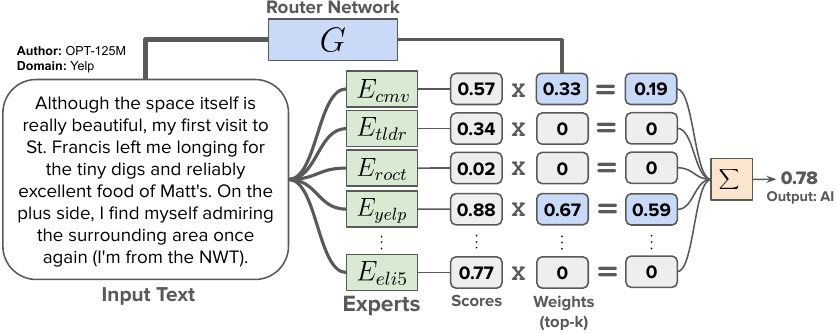}
    \caption{A graphical representation of the full Domain Gating Ensemble Network (DoGEN). Texts are given to each of $N$ domain-specific experts. Scores from each expert are ensembled using weights determined by the \textit{router network}. This network is trained for domain classification and the top-$k$ experts are chosen for output.}
    \label{fig:pg2}
\end{figure*}

\section{Related Work}
\paragraph{Neural Ensemble Methods} There have been many attempts to ensemble neural classifiers to detect machine-generated text. The most common approach collects zero-to-one scores from various classifiers and trains a lightweight classification layer on top to learn weights for the different classifiers \cite{abburi-etal-2023-simple,lai-etal-2024-adaptive,liyanage-buscaldi-2023-ensemble,nguyen-etal-2023-stacking,joy-aishi-2023-feature}. While this technique works well on in-domain data, it fails to generalize well to generators and domains that are not seen during training.

\paragraph{Metric-Based Ensembles} Another related line of work is ensembling features not derived from trained classifiers but rather derived from linguistic stylometry or logits of other LLMs. \citet{verma-etal-2024-ghostbuster} learns an ensemble by conducting a structured search over combinations of such features.  
\citet{ong2024applyingensemblemethodsmodelagnostic} ensemble the negative curvature of log likelihood from many different LLMs. Finally, \citet{mao-etal-2025-mlsdet} learn ensemble weights over log-rank and entropy from many different LLMs. While these techniques are effective at adapting to out-of-domain settings, they struggle to achieve high performance on in-domain text as their underlying features are generic and not optimized for any particular generator or domain.

\paragraph{Domain-Adaptation Techniques} Recent work has begun to incorporate techniques for learning domain-agnostic features for AI-text detection \cite{agrahari-etal-2025-random}. \citet{bhattacharjee-etal-2023-conda,bhattacharjee2024eagledomaingeneralizationframework} use Gradient Reversal Layers (GRL) along with a reconstruction loss. \citet{abassy-etal-2024-llm} use the same technique along with a multi-class classification objective. \citet{gui-etal-2025-aider} additionally learn a Variational Information Bottleneck (VIB) to assist in reconstruction. These techniques have shown a lot of promise, however, there are many valuable domain-specific features that one may want to use for detection. Our proposed method learns when---and when not---to use such features allowing for a more flexible approach.

\section{Our Approach}
Let $\mathcal{X}$ denote the space of input documents and $\mathcal{Y}$ the space of (scalar) detector scores
we wish to predict.\footnote{In our setting
$\mathcal{Y}=[0,1]$, where values closer to 1 indicate a higher probability that the text was
machine-generated.}
We fine-tune $N$ specialists (``experts’’) 
\[
E_i:\;\mathcal{X}\longrightarrow \mathcal{Y},\qquad i\in\{1,\dots,N\},
\]
each optimized on a single domain
$\mathcal{D}_i\subset\mathcal{X}$.
We then fine tune a router for domain classification $G:\mathcal{X}\!\to\!\Delta^{N-1}$ that maps each document to a
probability distribution over the $N$ domains $G(x;\phi, W)=(p_1,\dots,p_N)$,  
\[
p_i(x)=\frac{\exp\bigl(\mathbf{w}_i^{\top}\,\phi(x)\bigr)}
             {\sum_{j=1}^{N}\exp\bigl(\mathbf{w}_j^{\top}\,\phi(x)\bigr)},
\]
where $\phi(x)\in\mathbb{R}^d$ is an encoder representation and
$\{\mathbf{w}_i\}_{i=1}^{N}$ are learnable parameters.
During \emph{router training} we freeze the experts and minimize a standard
cross-entropy loss
\[
\mathcal{L}_{\text{gate}}
    =-\frac1M\sum_{m=1}^{M}\log p_{d_m}\!\bigl(x^{(m)}\bigr),
\]
where $d_m$ is the oracle domain label for sample $x^{(m)}$.
This loss encourages $G$ to give high likelihood to the domain–appropriate expert. Both the classification head $\{\mathbf{w}_i\}_{i=1}^{N}$ and the encoder $\phi$ are trained end-to-end.

\medskip
\noindent\textbf{Prediction rule.}
Given $p_1,\dots,p_N$ and the experts’ raw scores
$\mathbf{y}(x)=E_1(x),\dots,E_N(x)$ we compute the final detector score
with the following strategy: Let $\mathcal{I}_k$ be the indices of the $k$ largest $p_i(x)$,
      and softmax $w_i=p_i/\sum_{j\in\mathcal{I}_k}p_j$ for $i\in\mathcal{I}_k$.
      \[
      s^{(k)}(x)=
      \sum_{i\in\mathcal{I}_k} w_i \,E_i(x),\hspace{0.3em}
      k\in\{1,\dots,N\}
      \]
Setting $k=N$ recovers a full dot-product gating.

\section{Experiments}
\subsection{Data}
For our training data, we use the train split of \textbf{MAGE} \cite{li-etal-2024-mage}, a large corpus designed for training and evaluating machine-generated text detectors. The training set of MAGE contains text from 10 domains: \texttt{cmv}, \texttt{eli5}, \texttt{tldr}, \texttt{xsum}, \texttt{wp}, \texttt{roct}, \texttt{hswag}, \texttt{yelp}, \texttt{squad}, and \texttt{sci\_gen}. To address class imbalance, we down-sample the data to ensure an equal number of machine-generated and human-written documents per domain\footnote{see Appendix~\ref{sec:data-details} for ablations}. After down-sampling, each domain is split 90:10 into training and validation sets.

To test our models we use the test set of \textbf{MAGE} which includes data from these same 10 domains as well as 4 additional domains: \texttt{cnn}, \texttt{dialog\_sum}, \texttt{imdb}, and \texttt{pubmed}—which are unseen during training. In addition, we use the test set of the \textbf{RAID} benchmark \cite{dugan-etal-2024-raid}, focusing only on its non-adversarial subset to evaluate generalization across models, domains, and decoding strategies.

\subsection{Metrics}
For evaluation we use AUROC. This metric measures the probability that a detector ranks a randomly chosen machine-generated document higher than a human-written one. Because it integrates over all possible thresholds, AUROC is threshold-free, insensitive to class imbalance, and reflects overall ranking quality.

\subsection{Expert and Router Models (DoGEN)}
For DoGEN we fine-tune $N=10$ expert models for AI-text detection, one on each domain in the MAGE training set. Each expert uses a Qwen1.5-1.8B\footnote{\url{https://huggingface.co/Qwen/Qwen1.5-1.8B}} base model with a binary classification head. 

For the router network we again fine-tune Qwen1.5-1.8B to classify documents into the 10 domains in the training set of MAGE. At inference time we employ a top-$k$ routing strategy with $k=2$, meaning that each input is routed to the top two experts based on the router’s selection scores.

We fine-tune all models using the HuggingFace \texttt{Trainer} API. For full training details and hyperparameter settings, refer to Appendix~\ref{sec:hyperparameters}.

\subsection{Comparisons}
\label{sec:comparisons}
We compare DoGEN against the following baselines, all fine-tuned with the same hyper-parameters (Table~\ref{tab:train_hparams}) and class-balancing strategy:

\paragraph{Qwen1.5} As an initial baseline we trained two base models, Qwen1.5-1.8B and Qwen1.5-32B\footnote{\url{https://huggingface.co/Qwen/Qwen1.5-32B}} on all domains of MAGE. The 1.8B model was chosen to match the parameter count of a single expert and the 32B model was chosen to roughly match the parameter count of the full ensemble. This serves as a direct comparison to our experts, which all are fine-tuned versions of Qwen1.5. In essence, this helps us to understand how well an expert that was trained on all of MAGE would do on this task.

\begin{table*}[t]
\centering
\small
\setlength{\tabcolsep}{3pt}
\begin{tabular}{lcccccccccc|cccc|c}
\toprule
\multicolumn{16}{c}{\textbf{MAGE – AUROC (\%)}} \\ \midrule
\textbf{Model} & cmv & eli5 & tldr & xsum & wp & roct & hswag & yelp & squad & s\_gen & cnn & d\_sum & imdb & pubmed & all \\ \midrule
Qwen1.8B      & \textbf{99.89} & 98.33 & 98.49 & \textbf{99.71} & 98.97 & 98.89 & 98.89 & \textbf{98.69} & 97.79 & 97.74 & 71.58 & 39.05 & 91.56 & 81.97 & 88.41 \\
Qwen32B       & 98.52 & 96.22 & 94.71 & 96.60 & 96.62 & 91.06 & 91.03 & 95.15 & 97.91 & 97.94 & \textbf{93.52} & \textbf{90.50} & 93.89 & \textbf{90.97} & 95.20 \\
QwenMoE       & 98.60 & 95.91 & 96.35 & 90.10 & 98.87 & 94.78 & 94.97 & 95.17 & 98.03 & 94.08 & 86.53 & 75.11 & \textbf{96.08} & 87.66 & 95.43 \\
Equal Vt.      & 95.06 & 97.30 & 98.14 & 95.63 & 99.38 & 97.01 & 95.64 & 96.27 & 98.82 & 98.25 & 53.79 & 38.04 & 86.01 & 84.24 & 95.88 \\
Weight Vt. & 99.04 & 97.06 & 98.08 & 98.17 & 99.48 & 98.30 & 98.10 & 95.63 & 98.59 & 97.19 & 65.43 & 39.40 & 82.32 & 80.11 & 91.71 \\
JT-Domain & 99.74 & 98.50 & \textbf{98.99} & 99.09 & 99.56 & \textbf{98.92} & 98.08 & 98.20 & \textbf{99.43} & \textbf{99.28} & 61.44 & 62.49 & 81.62 & 77.38 & 96.74 \\
JT-Scratch & 98.91 & 96.51 & 96.09 & 96.21 & 99.20 & 91.76 & 89.30 & 94.68 & 98.01 & 98.56 & 76.59 & 62.28 & 81.23 & 80.26 & 94.17 \\
\midrule
\textbf{DoGEN} & 99.73 & \textbf{98.76} & \textbf{98.99} & 99.01 & \textbf{99.76} & 98.76 & \textbf{99.04} & 97.81 & 99.06 & 99.15 & 78.98 & 45.64 & 84.93 & 79.10 & \textbf{97.60} \\
\bottomrule
\end{tabular}
\caption{Performance on detecting machine-generated text in various domains from the MAGE benchmark. Bold values denote the best score in each column. We see that DoGEN performs the best overall on \textit{in-domain} data.}
\label{tab:mage_full}
\end{table*}

\begin{table*}[t]
\centering
\small
\begin{tabular}{l|ccccccccc}
\toprule
\multicolumn{10}{c}{\textbf{RAID – AUROC (\%)}} \\
\midrule
\textbf{Model} & abstracts & books & news & poetry & recipes & reddit & reviews & wiki & all \\ \midrule
Qwen1.8B & 92.95 & 94.93 & 95.95 & 91.18 & 84.30 & 91.33 & 96.52 & 90.64 & 93.34 \\
Qwen32B & 94.09 & 97.10 & 95.72 & 94.42 & 96.64 & 92.68 & 97.41 & 93.08 & 94.73 \\
QwenMoE & 88.24 & 91.99 & 92.81 & 92.75 & 82.22 & 93.75 & 94.72 & 95.75 & 91.34 \\
Equal Vote & 95.51 & \textbf{98.47} & 96.54 & 93.92 & 83.68 & \textbf{96.85} & \textbf{98.02} & 94.72 & 92.73 \\
Weighted Vote & 92.31 & 95.51 & 94.57 & 89.35 & 90.04 & 90.59 & 93.01 & 94.48 & 92.03 \\
JT-Domain & \textbf{96.84} & 96.73 & 97.86 & 90.85 & 69.89 & 94.73 & 97.10 & 96.34 & 92.27 \\
JT-Scratch & 95.36 & 97.10 & \textbf{98.51} & \textbf{98.08} & 87.49 & 94.57 & 94.47 & 96.37 & 95.65 \\
\midrule
\textbf{DoGEN} & 95.91 & 97.22 & 98.09 & 91.05 & \textbf{98.05} & 95.08 & 97.22 & \textbf{96.71} & \textbf{95.81} \\
\bottomrule
\end{tabular}
\caption{Performance of various models on detecting machine-generated text on the RAID benchmark. Bold values denote the best score in each column. We see that DoGEN performs the best overall on \textit{out-of-domain} data.}
\label{tab:raid_results_auroc}
\end{table*}
    
\paragraph{Qwen1.5-MoE-A2.7B}  This model adopts a Mixture-of-Experts (MoE) architecture by replacing the feed-forward layers in Qwen1.5-1.8B with MoE layers \cite{qwen_moe}. It has 14.3B parameters in total, with only 2.7B activated during runtime. We trained this model on all domains of MAGE and chose it model to highlight the specific advantage of our domain-aware gating strategy relative to a Mixture-of-Experts design.

\paragraph{Equal Vote} This comparison takes the trained experts and ensembles them together using an equal voting strategy. More specifically, given the experts’ raw scores
$\mathbf{y}(x)=E_1(x),\dots,E_N(x)$ we compute the final detector score as a weighted sum where all ensemble weights are $1/N$:
$\frac{1}{N}\sum_{j=1}^{N} E_j$.

\paragraph{Weighted Vote} This comparison follows \citet{abburi-etal-2023-simple} and uses Logistic Regression to learn ensemble weights for the $N$ experts. Unlike DoGEN, these weights are static and do not change based on the input text. The exact learned weights can be found in Appendix \ref{sec:ensemble-weights}.

\paragraph{Joint Training} This comparison takes both the router network $G$ and experts $E_{1}\cdots E_{N}$ and trains the full ensemble end-to-end on all domains. We conduct this experiment with two different initialization mechanisms. The first is initialized from the pre-trained Qwen1.5-1.8B model (JT-Scratch) and the second is initialized from the existing DoGEN model checkpoints (JT-Domain). Training was done with $k=N$ routing (all experts active) and evaluation was done with $k=2$ routing. We chose these comparisons to see whether the full ensemble would learn domain routing when trained end-to-end.

\section{Results}
\paragraph{In-domain (MAGE).}  
Table~\ref{tab:mage_full} reports performance on in-domain text from the MAGE benchmark. DoGEN achieves the highest overall AUROC (\textbf{97.60}\%), outperforming both the 32B dense baseline (95.20\%) and the off-the-shelf Mixture-of-Experts model (95.43\%). While the larger Qwen1.5-32B model performs well, it cannot specialize as effectively as our ensemble, which selects an expert fine-tuned for each specific domain.

At inference time, DoGEN activates only two 1.8B experts plus the gating network (5.4B parameters total), offering a parameter-efficient alternative to a 32B dense model. While the ensemble shows strong in-domain performance, it underperforms on certain out-of-distribution domains such as dialog\_sum (see Appendix~\ref{sec:dialogsum-challenges} for discussion). Nonetheless, DoGEN's modular structure allows us to add additional experts for targeted adaptation without retraining the entire model.

\paragraph{Out-of-domain (RAID).}  
Table~\ref{tab:raid_results_auroc} evaluates the same models on out-of-domain text from the RAID benchmark. 
We see that DoGEN again achieves the highest AUROC (95.81\%), outperforming both the 32B model (94.73\%) and the MoE baseline (91.43\%). Surprisingly, although the ensemble was designed for domain-specific specialization, it is able to generalize comparatively well to out-of-distribution examples. One possible explanation is that, on new unseen domains, the router network is able to choose experts that specialize in similar enough domains to still be able to predict accurately. We investigate this further in Appendix \ref{sec:expert-analysis}.

\section{Conclusion}
In the real-world, AI-generated text detection will mainly be used in a limited selection of high-risk domains (e.g. student essays, academic papers, news articles, etc.). In order to accurately detect text from these domains while remaining competitive in other rarer domains, detectors must be able to decide what models to use depending on the characteristics of the input text.

In this paper we propose a method to accomplish this by ensembling together a set of expert detectors via a gating network trained for domain classification. We show that not only do we achieve high accuracy on domains the model has seen during training, we also get very high performance on new, unseen domains---demonstrating the flexibility of the technique. 

\section*{Limitations}
One limitation of this paper is the lack of diversity of base models used in testing due to resource constraints. Our domain router and experts both use Qwen1.5-1.8B as a base and we do not test other potential models. Such experiments would allow us to rule out the possibility that the specific pre-training data distribution used to train Qwen can account for our positive result.

While we do test on many OOD domains and models, there are many other possible evaluation datasets that we did not choose to use such as SemEval \cite{semeval2024task8}, M4 \cite{wang-etal-2024-m4}, MULTITuDE \cite{macko-etal-2023-multitude}, etc. Future work should seek to incorporate multilingual text into this method---training a multilingual ensemble of experts with a language ID classification head is a promising future direction.

\section*{Acknowledgements}
This research is supported in part by the Office of the Director of National Intelligence (ODNI), Intelligence Advanced Research Projects Activity (IARPA), via the HIATUS Program contract \#2022-22072200005. The views and conclusions contained herein are those of the authors and should not be interpreted as necessarily representing the official policies, either expressed or implied, of ODNI, IARPA, or the U.S. Government. The U.S. Government is authorized to reproduce and distribute reprints for governmental purposes notwithstanding any copyright annotation therein.

\bibliography{custom}

\appendix

\section{Training Details}
\subsection{Hyperparameters \& Configurations}
\label{sec:hyperparameters}
\paragraph{Hyperparameters}
Table~\ref{tab:train_hparams} lists the full set of hyper-parameters used for all
experiments reported in this paper. The experts and domain routing network were trained using the Huggingface Trainer module while the Joint Training experiments were conducted using PyTorch. 

\begin{table}
\centering
\setlength{\tabcolsep}{5pt}
\renewcommand{\arraystretch}{1.05}
\begin{tabular}{@{}l c@{}}
\toprule
\textbf{Hyper‑parameter} & \textbf{Value} \\ \midrule
per\_device\_train\_batch\_size & 8 \\
per\_device\_eval\_batch\_size  & 8 \\
num\_train\_epochs             & 3 \\
learning\_rate                 & $5\times10^{-6}$ \\
evaluation\_strategy           & steps \\
eval\_steps                    & 100 \\
early\_stopping\_patience      & 10 \\
logging\_steps                 & 100 \\
dataloader\_num\_workers       & 2 \\
\bottomrule
\end{tabular}
\caption{Hyper‑parameters used in all experiments.}
\label{tab:train_hparams}
\end{table}

\subsection{Training Data Pre-processing}
\label{sec:data-details}

\begin{table*}[t]
\centering
\small
\resizebox{\textwidth}{!}{%
\begin{tabular}{l|cccccccccc|cccc|c}
\toprule
\multicolumn{16}{c}{\textbf{MAGE – AUROC (\%)}} \\
\midrule
\textbf{Strategy} & \texttt{cmv} & \texttt{eli5} & \texttt{tldr} & \texttt{xsum} & \texttt{wp} & \texttt{roct} & \texttt{hswag} & \texttt{yelp} & \texttt{squad} & \texttt{scigen} & \texttt{cnn} & \texttt{d\_sum} & \texttt{imdb} & \texttt{pubmed} & \textbf{All} \\
\midrule
Per-domain & \textbf{99.89} & \textbf{98.33} & \textbf{98.49} & \textbf{99.71} & 98.97 & \textbf{98.89} & \textbf{98.89} & \textbf{98.69} & \textbf{97.79} & \textbf{97.74} & \textbf{71.58} & \textbf{39.05} & \textbf{91.56} & \textbf{81.97} & \textbf{88.41} \\
Global & 91.77 & 80.41 & 75.76 & 79.78 & 92.51 & 75.24 & 56.01 & 40.64 & 65.71 & 76.01 & 25.83 & 38.17 & 79.33 & 60.00 & 71.98 \\
Unbalanced & 99.05 & 89.51 & 91.44 & 90.26 & \textbf{99.55} & 82.24 & 84.16 & 88.38 & 93.23 & 96.76 & 24.17 & 29.83 & 41.50 & 45.00 & 86.73 \\
\bottomrule
\end{tabular}
}
\caption{AUROC (\%) on MAGE-test of a single Qwen1.5-1.8B classifier trained on all documents in MAGE-train with various balancing strategies. \textbf{Per-domain}: per-domain training with human/AI balance. \textbf{Global}: Global balance across all domains. \textbf{Unbalanced}: No balancing done, Bold denotes the best score per column.}
\label{tab:mage_results_auroc}
\end{table*}

The distribution of human and AI-generated examples in the MAGE training data is highly imbalanced across domains prior to any preprocessing. For example, \texttt{roct} includes 25,510 AI examples but only 3,287 human examples, while \texttt{yelp} contains 31,827 human examples compared to 20,388 AI. In Table \ref{tab:data_counts} we report the original counts as well as the balanced counts used during training.

\begin{table}[t]
\centering
\setlength{\tabcolsep}{5pt}
\small
\begin{tabular}{l|c|c}
\toprule
\textbf{Domain} & \textbf{MAGE (Hum. / AI)} & \textbf{Balanced (Hum. / AI)} \\
\midrule
\texttt{cmv}       & 4,223 / 20,388 & 4,223 / 4,223 \\
\texttt{eli5}      & 16,706 / 25,548 & 16,706 / 16,706 \\
\texttt{hswag}     & 3,129 / 24,482 & 3,129 / 3,129 \\
\texttt{roct}      & 3,287 / 25,510 & 3,287 / 3,287 \\
\texttt{sci\_gen}  & 4,436 / 18,691 & 4,436 / 4,436 \\
\texttt{squad}     & 15,820 / 19,940 & 15,820 / 15,820 \\
\texttt{tldr}      & 2,826 / 19,811 & 2,826 / 2,826 \\
\texttt{wp}        & 6,356 / 24,803 & 6,356 / 6,356 \\
\texttt{xsum}      & 4,708 / 26,051 & 4,708 / 4,708 \\
\texttt{yelp}      & 31,827 / 20,529 & 20,529 / 20,529 \\
\bottomrule
\end{tabular}
\caption{The counts of documents for each domain for human and AI text before (left) and after (right) balancing. We balance the data such that each domain has an exactly 50:50 split.}
\label{tab:data_counts}
\end{table}

We observe that the per-domain balancing strategy consistently outperforms both dataset-wide and unbalanced training across in-distribution and out-of-distribution domains (see Table~\ref{tab:mage_results_auroc}). As a result, we adopt per-domain balancing for all expert and baseline training throughout our experiments.

\subsection{Hardware \& Training Time}

All models were trained using NVIDIA RTX A6000 GPUs with full-precision training. Training the baseline Qwen1.5–1.8B model used approximately 9.4 GPU hours, while the larger Qwen32B baseline required around 404.9 GPU hours. The QwenMoE model required approximately 402.3 GPU hours and training the domain experts, each based on Qwen1.5–1.8B, involved 10 domains and cumulatively consumed 235.1 GPU hours. The Joint Training models JT-Scratch and JT-Domain took an additional 861 GPU hours each. In total, model training across all baselines, experts, and ensembles required 2,773.7 GPU hours.

\subsection{``Weighted Vote'' Ensemble Weights}
\label{sec:ensemble-weights}

For the ``Weighted Vote'' comparison we train a Logistic Regression classifier using \texttt{scikit-learn} on the 10-dimensional vector of expert scores. We use the
\texttt{lbfgs} solver, \texttt{max\_iter}=1000, and
\texttt{class\_weight=balanced}.  Prior to fitting, expert scores are
standardised with \texttt{StandardScaler}. Table~\ref{tab:ensemble-weights} reports the resulting normalised
coefficients; they sum to~1 such that larger values indicate greater
average influence on the final decision.

\begin{table}[t]
\centering
\small
\setlength{\tabcolsep}{6pt}
\renewcommand{\arraystretch}{1.05}
\begin{tabular}{lc}
\toprule
\textbf{Expert} & \textbf{Ensemble weight} \\
\midrule
xsum     & 0.181 \\
roct     & 0.141 \\
hswag    & 0.107 \\
eli5     & 0.107 \\
squad    & 0.094 \\
cmv      & 0.093 \\
wp       & 0.088 \\
tldr     & 0.078 \\
yelp     & 0.058 \\
sci\_gen & 0.054 \\
\bottomrule
\end{tabular}
\caption{Normalised logistic-regression coefficients
showing each specialist’s contribution to the ``Weighted Vote'' ensemble. Higher weights correspond to a stronger influence on the ensemble’s
output probability.}
\label{tab:ensemble-weights}
\end{table}

\begin{table*}[t]
\centering
\small
\begin{tabular}{l|l|l}
\toprule
\textbf{Domain} & \textbf{Source Dataset} & \textbf{Description} \\
\midrule
\textbf{CMV} & \href{https://dl.acm.org/doi/10.1145/2872427.2883081}{r/ChangeMyView} & Opinion statements from Reddit debates. \\
\textbf{Yelp} & \href{https://proceedings.neurips.cc/paper_files/paper/2015/file/250cf8b51c773f3f8dc8b4be867a9a02-Paper.pdf}{Yelp Reviews} & Restaurant and service reviews by users. \\
\textbf{XSum} & \href{https://aclanthology.org/D18-1206/}{BBC Summaries} & Short one-sentence news article summaries. \\
\textbf{TLDR} & \href{https://huggingface.co/datasets/JulesBelveze/TLDR_news}{TLDR\_news} & Concise news article summaries. \\
\textbf{ELI5} & \href{https://aclanthology.org/P19-1346/}{Explain Like I'm 5} & Long answers to simple crowd-asked questions. \\
\textbf{WP} & \href{https://aclanthology.org/P18-1082/}{r/WritingPrompts} & Creative stories from short prompts. \\
\textbf{ROC} & \href{https://aclanthology.org/N16-1098/}{ROCStories} & Commonsense, causal short stories. \\
\textbf{HellaSwag} & \href{https://aclanthology.org/P19-1472/}{HellaSwag} & Sentence completion for commonsense reasoning. \\
\textbf{SQuAD} & \href{https://aclanthology.org/D16-1264/}{SQuAD} & Wikipedia passages for QA context. \\
\textbf{SciGen} & \href{https://aclanthology.org/2021.findings-emnlp.128/}{SciXGen} & Scientific article abstracts and prompts. \\
\midrule
\textbf{CNN} & \href{https://aclanthology.org/P17-1099/}{CNN/DailyMail} & Articles paired with multi sentence summaries. \\
\textbf{DialogSum} & \href{https://aclanthology.org/2021.findings-acl.449/}{DialogSum} & Realistic everyday conversations and summaries. \\
\textbf{PubMedQA} & \href{https://aclanthology.org/D19-1259/}{PubMedQA} & QA pairs from medical research literature. \\
\textbf{IMDb} & \href{https://aclanthology.org/P11-1015/}{IMDb Reviews} & Sentiment-rich movie reviews. \\
\bottomrule
\end{tabular}
\caption{Sources and descriptions for the 10 train and 4 held-out test domains included in the MAGE dataset.}
\label{tab:mage_domains_extended}
\end{table*}

\begin{table}[t]
\centering
\small
\begin{tabular}{p{0.35\linewidth}|p{0.55\linewidth}} 
\toprule
\textbf{Model} & \textbf{Identifier} \\
\midrule
OpenAI GPT\newline  \cite{brown2020language} & \texttt{text-davinci-002}, \texttt{text-davinci-003}, \texttt{gpt-3.5-turbo} \\
LLaMA\newline \cite{touvron2023llama} & \texttt{LLaMA-6B}, \texttt{LLaMA-13B}\newline\texttt{LLaMA-30B}, \texttt{LLaMA-65B} \\
GLM \newline \cite{zeng2022glm130b} & \texttt{GLM-130B} \\
FLAN-T5\newline \cite{chung2022scaling} & \texttt{flan-t5-small}, \texttt{flan-t5-base}, \texttt{flan-t5-large}, \texttt{flan-t5-xl}, \texttt{flan-t5-xxl} \\
OPT\newline \cite{zhang2022opt} & \texttt{opt-125M}, \texttt{opt-350M}, \texttt{opt-1.3B}, \texttt{opt-2.7B}, \texttt{opt-6.7B}, \texttt{opt-13B}, \texttt{opt-30B}, \texttt{opt-iml-1.3B}, \texttt{opt-iml-30B} \\
BigScience\newline \cite{sanh2022multitask} & \texttt{T0-3B}, \texttt{T0-11B}, \texttt{BLOOM-7B1} \\
EleutherAI\newline \cite{black2022gptneox} & \texttt{GPT-J-6B}, \texttt{GPT-NeoX-20B} \\
\bottomrule
\end{tabular}
\caption{Full set of large language models used in the MAGE dataset. Each family is cited once and includes all released variants used in the dataset.}
\label{tab:mage_generators_full}
\end{table}

\begin{table}[t]
\centering 
\small
\begin{tabular}{l|l} 
\toprule
\textbf{Model}&\textbf{Identifier}\\
\midrule
GPT-2&\texttt{gpt2-xl}\\
\cite{radford2019language}&\\
MPT (\textit{+ Chat}) &\texttt{mpt-30b}\\
\cite{MosaicML2023Introducing}&\texttt{mpt-30b-chat}\\
Mistral (\textit{+ Chat})&\texttt{Mistral-7B-v0.1}\\
\cite{jiang2023mistral}&\texttt{Mistral-7B-Instruct-v0.1}\\
LLaMA Chat&\texttt{Llama-2-70b-chat-hf}\\
\cite{touvron2023llama}&\\
Cohere (\textit{+ Chat})&\texttt{command (co.generate())}\\
\cite{Cohere2024}&\texttt{command (co.chat())}\\
GPT-3&\texttt{text-davinci-002}\\
\cite{ouyang2022training}&\\
ChatGPT&\texttt{gpt-3.5-turbo-0613}\\
\cite{chatgpt-etal-2022}&\\
GPT-4&\texttt{gpt-4-0613}\\
\cite{openai2023gpt4}&\\
\bottomrule
\end{tabular}
\caption{The generative models used in the RAID dataset along with citations to the original papers.}
\label{tab:generative_models}
\end{table}

\begin{table}[t]
\centering 
\small
\begin{tabular}{p{0.17\linewidth}|p{0.28\linewidth}|p{0.34\linewidth}} 
\toprule
\textbf{Domain}&\textbf{Source}&\textbf{Description}\\
\midrule
\textbf{Abstracts}&\href{https://www.kaggle.com/datasets/Cornell-University/arxiv}{arxiv.org}&ArXiv Abstracts\\
\textbf{Recipes}&\href{https://recipenlg.cs.put.poznan.pl/}{allrecipes.com}&Ingredients + Recipe\\
\textbf{Books}&\href{https://paperswithcode.com/dataset/cmu-book-summary-dataset}{wikipedia.org}&Plot Summaries\\
\textbf{Reddit}&\href{https://huggingface.co/datasets/sentence-transformers/reddit-title-body}{reddit.com}&Reddit Posts\\
\textbf{News}&\href{https://github.com/derekgreene/bbc-datasets}{bbc.com/news}&News Articles\\
\textbf{Reviews}&\href{https://ieee-dataport.org/open-access/imdb-movie-reviews-dataset}{imbd.com}&Movie Reviews\\
\textbf{Poetry}&\href{https://www.kaggle.com/datasets/michaelarman/poemsdataset}{poemhunter.com}&Poems (Any Style)\\
\textbf{Wiki}&\href{https://huggingface.co/datasets/aadityaubhat/GPT-wiki-intro}{wikipedia.org}&Article Introductions\\
\bottomrule
\end{tabular}
\caption{All domains in the RAID dataset alongside a description of where they are from. Clickable source links go directly to the source dataset from which the human samples were taken.}
\label{tab:domains}
\end{table}

\section{Domains and Generators}
In this section we briefly discuss the different domains and generators present in the MAGE \cite{li-etal-2024-mage} and RAID \cite{dugan-etal-2024-raid} benchmarks. We do this to help readers better understand the similarities and differences between the two domains and to give extra context on the transferability of our approach. We recommend reading the original papers for more detailed descriptions.

\subsection{MAGE Domains}
In Table \ref{tab:mage_domains_extended} we report descriptions and sources for the 14 domains in MAGE. Of particular note is the DialogSum domain, which consists of natural conversations and is significantly different from other domains present in MAGE (see Appendix \ref{sec:dialogsum-challenges}). In general text was taken from publicly available sources such as Wikipedia, Reddit, Yelp, ArXiv, and News platforms (CNN, BBC, etc.).

\subsection{MAGE Generators}
\label{sec:mage_generators}
In Table \ref{tab:mage_generators_full} we report the full list of 27 models used to generate the training split of the MAGE dataset. We see that the MAGE dataset is primarily made up of models that are pre-ChatGPT such as Llama-1 and FLAN-T5. We also see that the models used to generate are mainly on the smaller end (125M-7B) as compared to the average parameter count of more recent models. This makes it all the more surprising that DoGEN generalizes well to the RAID benchmark as it primarily consists of text generated by models released after ChatGPT.

In addition to these models, the authors of MAGE use GPT-4 \cite{openai2023gpt4} to generate examples for the held out domains (CNN, DialogSum, PubMed, IMDb) and paraphrase both the GPT-4 examples and the human-written examples with GPT3.5 \cite{chatgpt-etal-2022}. This makes the OOD domains significantly different from the main MAGE train distribution (see Appendix \ref{sec:mage_ood}).

\subsection{RAID Domains}
In Table \ref{tab:generative_models} we list the domains present in the RAID dataset \cite{dugan-etal-2024-raid}. Many of the domains come from similar sources as MAGE. For example, the Wikipedia, Reddit, News, Abstracts, and Books domains all share common origins with domains present in MAGE. While this doesn't mean they are exactly the same distribution of text, it does means that we should expect ensembles trained on MAGE to perform relatively well on these splits.

The other domains that are dissimilar to MAGE are Reviews, Poetry, and Recipes. These are unusual sources for machine-generated text and, as expected, DoGEN does comparatively worse on these domains (see Table \ref{tab:raid_results_auroc}).

\subsection{RAID Generators}
In Table \ref{tab:generative_models} we describe the models used to generate the RAID dataset. We see that most models are from mid-to-late 2023 and are likely quite different from the models present in MAGE. 

Some similarities between the two sets of generators are present. For example, both datasets use text-davinci-002 and gpt-3.5-turbo. In addition, both datasets use a version of Llama for generation. However, RAID also includes models like Mistral, Cohere, and MPT, which are not included in either MAGE-train or test.

\section{Additional Expert Analysis}
\label{sec:expert-analysis}

Tables~\ref{tab:expert_mage} and~\ref{tab:expert_raid} show that each
specialist achieves its peak \emph{AUROC} on domains it was fine-tuned
for.  We now examine whether the router exploits this specialisation.

\paragraph{The router allocates weight to the most suitable specialists.}
For every RAID document \(x\) we record

\begin{itemize}\setlength\itemsep{2pt}
  \item the probability \(p_i(x)\) assigned to expert \(E_i\)
  \item the stand-alone AUROC \(a_i\) that \(E_i\) attains on the full
        RAID corpus (Table~\ref{tab:gate-corr}, col.~2).
\end{itemize}

\begin{figure}[t]
  \centering
  \includegraphics[width=\linewidth]{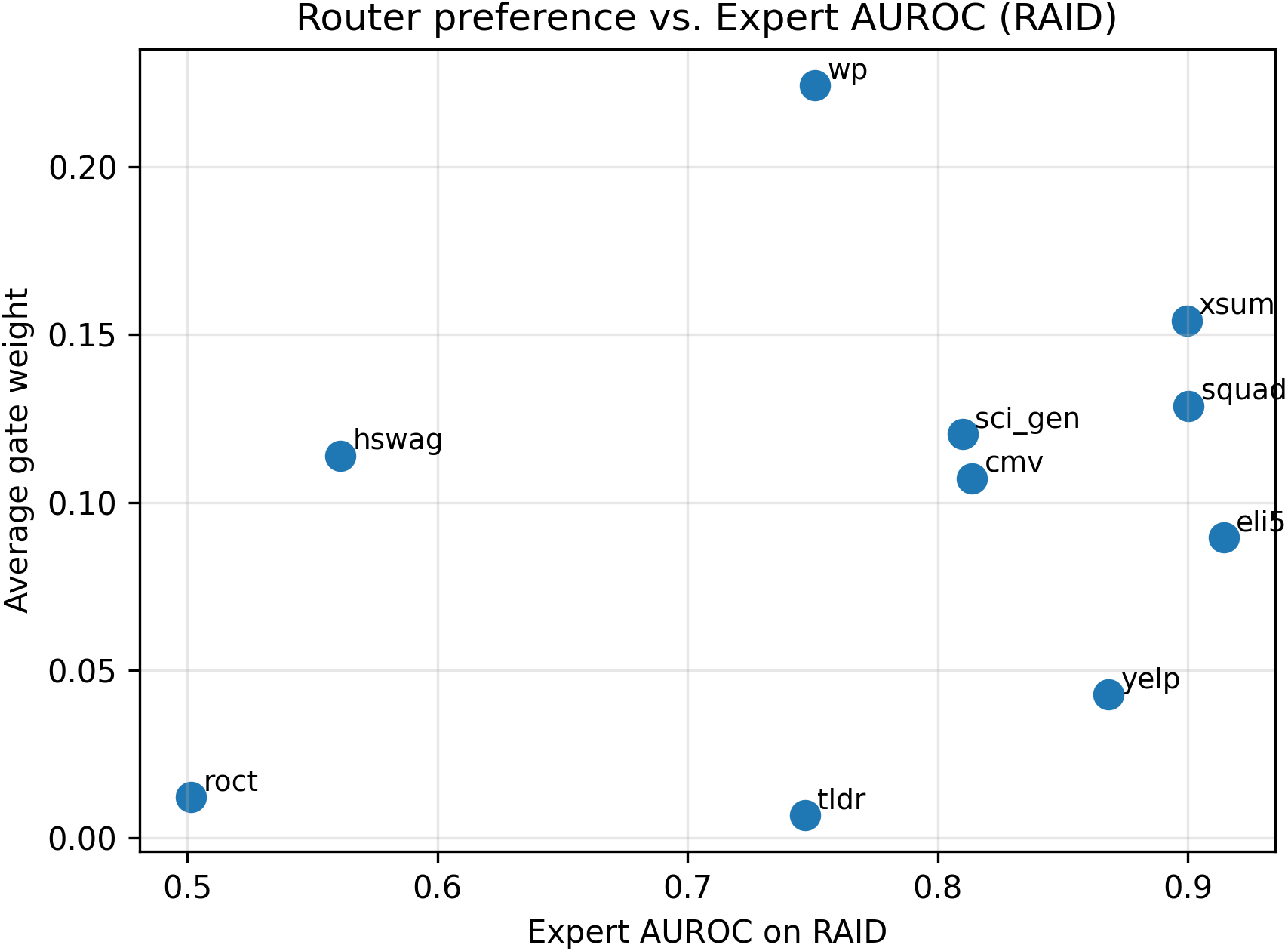}
  \caption{Average gate weight versus expert AUROC on RAID.
           Pearson \(\rho = 0.64,\;p < 0.03\).}
  \label{fig:gate_vs_auc}
\end{figure}

A positive association would indicate that the router systematically
favours those specialists whose decision boundaries generalise best to
the current input distribution.  
Figure~\ref{fig:gate_vs_auc} confirms this intuition
(Pearson \(\rho = 0.64,\;p < 0.03\)): experts with larger AUROC attract more
probability mass.

\begin{table}
\centering
\small
\setlength{\tabcolsep}{5pt}
\renewcommand{\arraystretch}{1.05}
\begin{tabular}{lccc}
\toprule
\textbf{Expert} & \textbf{AUROC} & \textbf{Avg.\,$\bar p_i$} & \boldmath{$r_i$} \\
\midrule
eli5      & 0.914 & 0.22 & 0.45 \\
squad     & 0.846 & 0.19 & 0.40 \\
cmv       & 0.814 & 0.17 & 0.35 \\
sci\_gen  & 0.810 & 0.16 & 0.32 \\
yelp      & 0.804 & 0.15 & 0.30 \\
xsum      & 0.760 & 0.12 & 0.25 \\
tldr      & 0.742 & 0.10 & 0.20 \\
wp        & 0.739 & 0.08 & 0.15 \\
hswag     & 0.561 & 0.05 & 0.08 \\
roct      & 0.501 & 0.03 & 0.05 \\
\bottomrule
\end{tabular}
\caption{Standalone AUROC on RAID (``all'' column), average gate weight
$\bar p_i$, and per-expert Pearson correlation
$r_i=\mathrm{corr}\!\bigl(p_i,\mathbf 1\{\text{correct}\}\bigr)$.
Higher-AUROC specialists receive larger weights and exhibit stronger
positive correlations.}
\label{tab:gate-corr}
\end{table}

\paragraph{Take-away.}
Experts are neither intrinsically “good’’ nor “bad’’—they are specialised
to distinct domains.  
By first inferring the domain of each document, the router \emph{learns
to trust a specialist exactly when that specialist’s domain matches the
input} and to down-weight specialists that are out-of-distribution for
the current example.  This domain-aware routing allows the ensemble to
generalise robustly to unseen genres without explicit supervision or
hand-crafted rules.

\begin{table*}[t]
\centering
\setlength{\tabcolsep}{3pt}
\renewcommand{\arraystretch}{1.1}
\small

\begin{tabular}{@{}lcccccccccc|cccc|c@{}}
\toprule
\multicolumn{1}{c}{} & \multicolumn{10}{c|}{\textbf{MAGE (In-Domain)}} & \multicolumn{4}{c|}{\textbf{MAGE (Out-of-Domain)}} & \\
\textbf{Expert} & cmv & eli5 & tldr & xsum & wp & roct & hswag & yelp & squad & sci\_gen & cnn & d\_sum & imdb & pubmed & all \\
\midrule
cmv      & \textbf{0.998} & 0.955 & 0.741 & 0.677 & 0.980 & 0.806 & 0.797 & 0.909 & 0.844 & 0.866 & 0.831 & \textbf{0.970} & 0.924 & 0.550 & 0.831 \\
eli5      & 0.963 & \textbf{0.989} & 0.943 & 0.716 & 0.978 & 0.905 & 0.910 & 0.946 & 0.954 & 0.913 & \textbf{0.897} & 0.956 & \textbf{0.977} & \textbf{0.753} & \textbf{0.897} \\
tldr    & 0.787 & 0.787 & \textbf{0.995} & 0.751 & 0.810 & 0.830 & 0.812 & 0.838 & 0.832 & 0.869 & 0.721 & 0.830 & 0.841 & 0.603 & 0.799 \\
xsum     & 0.796 & 0.798 & 0.838 & \textbf{0.995} & 0.856 & 0.742 & 0.749 & 0.809 & 0.764 & 0.832 & 0.672 & 0.859 & 0.834 & 0.605 & 0.770 \\
wp        & 0.875 & 0.842 & 0.772 & 0.688 & \textbf{0.997} & 0.765 & 0.772 & 0.810 & 0.800 & 0.850 & 0.732 & 0.840 & 0.818 & 0.602 & 0.781 \\
roct     & 0.444 & 0.579 & 0.798 & 0.427 & 0.507 & \textbf{0.989} & 0.846 & 0.577 & 0.642 & 0.688 & 0.654 & 0.488 & 0.487 & 0.506 & 0.654 \\
hswag    & 0.457 & 0.614 & 0.678 & 0.400 & 0.430 & 0.564 & \textbf{0.992} & 0.593 & 0.709 & 0.697 & 0.608 & 0.962 & 0.498 & 0.488 & 0.608 \\
yelp     & 0.842 & 0.871 & 0.867 & 0.744 & 0.890 & 0.813 & 0.799 & \textbf{0.996} & 0.851 & 0.904 & 0.736 & 0.913 & 0.866 & 0.671 & 0.814 \\
squad    & 0.946 & 0.932 & 0.899 & 0.781 & 0.948 & 0.776 & 0.786 & 0.933 & \textbf{0.982} & 0.899 & 0.840 & 0.935 & 0.917 & 0.717 & 0.857 \\
sci\_gen & 0.905 & 0.790 & 0.863 & 0.746 & 0.881 & 0.749 & 0.833 & 0.846 & 0.907 & \textbf{0.971} & 0.803 & 0.962 & 0.846 & 0.636 & 0.803 \\
\bottomrule
\end{tabular}
\caption{Expert-Level AUROC Results on MAGE. Columns grouped into In-Domain and Out-of-Domain (OOD) Best performance in each column is bolded. d\_sum is shorthand for dialog\_sum}
\label{tab:expert_mage}
\end{table*}

\begin{table*}[t]
\centering
\setlength{\tabcolsep}{4pt}
\renewcommand{\arraystretch}{1.1}
\small

\begin{tabular}{@{}lcccccccc|c@{}}
\toprule
\multicolumn{10}{c}{\textbf{RAID (Out-of-Domain) – AUROC}} \\ \midrule
\textbf{Expert} & abstracts & books & news & poetry & recipes & reddit & reviews & wiki & \textbf{all} \\
\midrule
cmv\_expert       & \textbf{0.970} & 0.924 & 0.550 & 0.896 & 0.796 & 0.931 & 0.946 & 0.580 & 0.814 \\
eli5\_expert      & 0.956 & \textbf{0.977} & \textbf{0.753} & \textbf{0.943} & \textbf{0.939} & \textbf{0.948} & \textbf{0.973} & \textbf{0.861} & \textbf{0.914} \\
tldr\_expert      & 0.804 & 0.791 & 0.644 & 0.702 & 0.739 & 0.760 & 0.778 & 0.719 & 0.742 \\
xsum\_expert      & 0.812 & 0.801 & 0.676 & 0.747 & 0.765 & 0.760 & 0.784 & 0.729 & 0.760 \\
wp\_expert        & 0.834 & 0.782 & 0.639 & 0.697 & 0.743 & 0.755 & 0.776 & 0.690 & 0.739 \\
roct\_expert      & 0.488 & 0.487 & 0.506 & 0.535 & 0.489 & 0.493 & 0.533 & 0.480 & 0.501 \\
hswag\_expert     & 0.962 & 0.498 & 0.488 & 0.533 & 0.496 & 0.492 & 0.524 & 0.504 & 0.561 \\
yelp\_expert      & 0.876 & 0.864 & 0.697 & 0.782 & 0.805 & 0.815 & 0.838 & 0.753 & 0.804 \\
squad\_expert     & 0.905 & 0.854 & 0.730 & 0.791 & 0.864 & 0.870 & 0.905 & 0.852 & 0.846 \\
sci\_gen\_expert  & 0.962 & 0.846 & 0.636 & 0.610 & 0.895 & 0.776 & 0.898 & 0.828 & 0.810 \\
\bottomrule
\end{tabular}
\caption{Expert-Level AUROC Results on RAID (Out-of-Domain). Best performance in each column is bolded.}
\label{tab:expert_raid}
\end{table*}

\section{Understanding Poor MAGE Out-of-Domain Performance}
\label{sec:mage_ood}

\begin{table*}[t]
\setlength{\tabcolsep}{3pt}
\centering
\small
\begin{tabular}{@{}l|cccc|cccc|cccc|cccc@{}}
\toprule
& \multicolumn{4}{c|}{\textbf{CNN}} & \multicolumn{4}{c|}{\textbf{DialogSum}} & \multicolumn{4}{c|}{\textbf{IMDb}} & \multicolumn{4}{c}{\textbf{PubMedQA}} \\
& gpt4 & gpt4-p & hum-p & all & gpt4 & gpt4-p & hum-p & all& gpt4 & gpt4-p & hum-p & all& gpt4 & gpt4-p & hum-p & all \\
\midrule
Qwen1.8B & 71.6 & 66.7 & 54.0 & 71.6 & 61.8 & 31.9 & 43.4 & 39.1 & 91.5 & 95.9 & 79.3 & 91.6 & 82.0 & 88.1 & 58.1 & 82.0 \\
Qwen32B & \textbf{98.6} & \textbf{95.9} & \textbf{86.1} & \textbf{93.5} & \textbf{94.8} & \textbf{91.7} & \textbf{85.0} & \textbf{90.5} & 97.7 & \underline{96.7} & \underline{87.4} & \underline{93.9} & 96.1 & \textbf{93.4} & \textbf{83.5} & \textbf{91.0} \\
QwenMoE & 91.2 & \underline{89.6} & \underline{78.8} & \underline{86.5} & 91.0 & \underline{80.5} & \underline{53.8} & \underline{75.1} & \textbf{98.9} & \textbf{97.7} & \textbf{91.6} & \textbf{96.1} & 98.6 & 89.5 & 74.9 & 87.7 \\
Equal Vt.& 87.3 & 37.9 & 36.2 & 53.8 & \underline{91.6} & 37.1 & 25.4 & 38.0 & 95.8 & 90.6 & 71.6 & 86.0 & \textbf{99.2} & 90.2 & 63.3 & 84.2 \\
Weight Vt.& 91.5 & 64.5 & 40.3 & 65.4 & 82.7 & 38.8 & 26.8 & 39.4 & 96.1 & 88.5 & 71.3 & 82.3 & \underline{99.0} & 84.5 & 61.8 & 80.1 \\
JT-Domain& 86.2 & 53.0 & 45.1 & 61.6 & 86.6 & 39.8 & 51.1 & 62.5 & \underline{98.3} & \underline{96.7} & 79.9 & 91.6 & 94.7 & 79.9 & 57.5 & 77.4 \\
JT-Scratch& 90.7 & 70.6 & 68.5 & 76.6 & 85.4 & 62.2 & 39.3 & 62.3 & 95.3 & 90.8 & 78.5 & 81.2 & 97.7 & 83.8 & 62.4 & 80.3 \\
\midrule
\textbf{DoGEN} & \underline{97.1} & 85.5 & 78.3 & 78.9 & 73.3 & 51.5 & 32.1 & 45.6 & 95.3 & 93.2 & 78.3 & 84.9 & 97.2 & 84.2 & \underline{78.9} & 79.1 \\
\bottomrule
\end{tabular}
\caption{AUROC (\%) scores for MAGE Out-of-Domain. ``gpt4-p'' and ``hum-p'' refer to gpt4 and human texts paraphrased by \texttt{gpt-3.5-turbo} (see Section \ref{sec:mage_generators}). We see that while DoGEN does poorly on aggregate metrics, on gpt4 generations it does well in every domain except dialog\_sum (see Section \ref{sec:dialogsum-challenges}).}
\label{tab:mage_nods}
\end{table*}

\subsection{Paraphrased Data}

The test domains in MAGE include text samples that differ meaningfully from the data used to train our expert models. While domain shift is expected in out-of-domain (OOD) evaluations, two of the three evaluation settings used in MAGE introduce particularly unusual cases.

\begin{table*}[t]
\centering
\small
\begin{tabular}{l|ccccccccc}
\toprule
\multicolumn{10}{c}{\textbf{RAID – TPR@FPR=5\%}} \\
\midrule
\textbf{Model} & abstracts & books & news & poetry & recipes & reddit & reviews & wiki & all \\
\midrule
Qwen1.8B & 85.01 & 84.44 & 84.19 & 82.59 & 74.22 & 81.79 & 85.26 & 83.56 & 82.63 \\
Qwen32B & \textbf{91.85} & 86.66 & 91.62 & 82.41 & \textbf{86.97} & 73.34 & 88.38 & 75.10 & 84.54 \\
QwenMoE & 72.93 & 90.46 & 84.25 & 79.03 & 82.84 & 85.65 & \textbf{93.44} & 87.10 & 84.46 \\
Equal Vote & 87.68 & 90.68 & 87.79 & 68.84 & 72.41 & \textbf{89.07} & 90.38 & 82.62 & 83.68 \\
Weighted Vote & 86.24 & 88.24 & 86.12 & 70.89 & 75.13 & 79.57 & 80.20 & 83.11 & 83.75 \\
JT-Domain & 90.63 & \textbf{91.65} & 91.56 & 80.94 & 21.41 & 86.69 & 88.82 & 84.34 & 79.51 \\
JT-Scratch & 89.63 & 89.10 & \textbf{95.28} & \textbf{92.40} & 39.70 & 85.60 & 81.96 & 87.01 & 82.58 \\
\midrule
\textbf{DoGEN} (ours) & 89.28 & 91.13 & 91.76 & 75.16 & 79.81 & 82.82 & 89.76 & \textbf{87.16} & \textbf{85.86} \\
\bottomrule
\end{tabular}
\caption{True Positive Rate at 5\% False Positive Rate (TPR@FPR=5\%) across RAID domains. Bold values indicate the best score in each column.}
\label{tab:raid_tpr_results}
\end{table*}

Specifically, the \texttt{gpt4\_para} setting includes outputs from GPT-4 that have been paraphrased by \texttt{gpt-3.5-turbo}. Likewise, the \texttt{human\_para} setting consists of human-written text that has been paraphrased by \texttt{gpt-3.5-turbo}. These test cases do not cleanly align with either "machine-generated" or "human-written" categories and introduce ambiguity into the labeling scheme.

As shown in Table~\ref{tab:mage_nods}, our model actually performs quite well at detecting \texttt{human\_para} text as human-written. This result supports our position that such examples—although technically modified by a machine—are stylistically closer to human writing and their inclusion in the "machine" class may be misleading. We argue that the use of paraphrased human text as a negative class label introduces noise into the evaluation and may inflate or deflate detector performance in ways that are difficult to interpret.

Despite this ambiguity, DoGEN performs competitively on \texttt{gpt4} examples—an especially relevant case, as GPT-4 was not seen by any expert during training. This suggests that DoGEN is capable of generalizing to unseen generators and adapting to challenging OOD distributions, performing comparably or better than other baselines.

\subsection{Challenges with the \texttt{dialogsum} Domain}
\label{sec:dialogsum-challenges}

One domain where our models notably underperform is \texttt{dialogsum}. This domain differs significantly in structure from our in-domain training sets. All of our in-domain examples (e.g., \texttt{cmv}, \texttt{eli5}, \texttt{xsum}) are prose-style summaries or longform posts, typically formatted as block paragraphs of explanatory or narrative text. In contrast, \texttt{dialogsum} consists of multi-speaker conversational transcripts, with speaker tags, turn-taking, short utterances, and question–answer structure.

\begin{table}
\centering
\small
\begin{tabular}{p{0.9\columnwidth}}
\toprule
\textbf{Example from the \texttt{dialogsum} domain:} \\
\midrule
``Person1: Hello. Person2: Hello. May I speak to Mark, please? Person1: I apologize, but Mark is not currently available. Would you like me to take a message or assist you with something else?'' \\
\bottomrule
\end{tabular}
\caption{Quoted example from the \texttt{dialogsum} domain.}
\label{tab:dialogsum_single_quote}
\end{table}

Our experts were never trained on dialogue-style input. Consequently, the gating classifier has difficulty assigning dialogsum inputs to the correct expert, and the experts themselves struggle to map multi-turn dialogue into effective summaries. This mismatch leads to weaker performance on this domain. An example of such dialog-style input is shown in Table~\ref{tab:dialogsum_single_quote}.

\subsection{Adapting with DoGEN}
A key strength of the DoGEN architecture is its modularity. Because each expert is independently trainable, we can train a new expert specifically on \texttt{dialogsum} examples and retrain the domain classifier to incorporate it. This offers a principled and efficient way to extend the system to structurally novel inputs without the need to retrain all experts or overhaul the entire ensemble.

By addressing domain-specific structural mismatches and class imbalance, we ensure that our evaluation of generalization is both fair and adaptable to new domains.

\section{Additional Results}
\paragraph{RAID Results TPR@FPR=5\%}
For our evaluation on RAID we also report \textbf{True Positive Rate at a 5\% False Positive Rate} (TPR@FPR$=5\%$) in Table \ref{tab:raid_tpr_results}. This quantifies how effectively a detector identifies machine-generated text while limiting false positives on human-written content to 5\%. We compute this metric using threshold values obtained via the RAID benchmark API across all evaluation domains. We do this to match the more standard evaluation metric that is calculated for submission to the RAID benchmark allowing us to be comparable to other established detection methods.

\section{AI Assistance}
During the writing of this paper AI assistants were used to help enhance clarity of writing, format tables and figures, as well as draft language for the formalisms in the methods section. The assistants used were ChatGPT and OpenAI o3.

\end{document}